# T-GMSI: A transformer-based generative model for spatial interpolation under sparse measurements

Xiangxi Tian and Jie Shan
School of Civil Engineering, Purdue University
West Lafayette, IN 47907, USA
{tian133, jshan}@purdue.edu

**Abstract**

Generating continuous environmental models from sparsely sampled data is a critical challenge in spatial modeling, particularly for topography. Traditional spatial interpolation methods often struggle with handling sparse measurements. To address this, we propose a Transformer-based Generative Model for Spatial Interpolation (T-GMSI) using a vision transformer (ViT) architecture for digital elevation model (DEM) generation under sparse conditions. T-GMSI replaces traditional convolution-based methods with ViT for feature extraction and DEM interpolation while incorporating a terrain feature-aware loss function for enhanced accuracy. T-GMSI excels in producing high-quality elevation surfaces from datasets with over 70% sparsity and demonstrates strong transferability across diverse landscapes without fine-tuning. Its performance is validated through extensive experiments, outperforming traditional methods such as ordinary Kriging (OK) and natural neighbor (NN) and a conditional generative adversarial network (CGAN)-based model (CEDGAN). Compared to OK and NN, T-GMSI reduces root mean square error (RMSE) by 40% and 25% on airborne lidar data and by 23% and 10% on spaceborne lidar data. Against CEDGAN, T-GMSI achieves a 20% RMSE improvement on provided DEM data, requiring no fine-tuning. The ability of model on generalizing to large, unseen terrains underscores its transferability and potential applicability beyond topographic modeling. This research establishes T-GMSI as a state-of-the-art solution for spatial interpolation on sparse datasets and highlights its broader utility for other sparse data interpolation challenges.

# 1. Introduction

Due to the incapability of acquiring information at all desired locations, the need to maximize the available information from sparsely distributed data or measurements is especially pertinent within an applied environmental modeling context, such as the spatial distribution of topography, temperature, precipitation and other natural or man-made phenomena (Li and Heap, 2014, 2008; Mitas and Mitasova, 1999). When dealing with the environmental modeling of a large scale or extent, denser spatial data often exponentially increases the cost of data acquisition. Indeed, the sparsity of the sampled data often limits the resolution of the final continuous product. For example, two recent and ongoing NASA's space missions, the Ice, Cloud, and land Elevation Satellite-2 (ICESat-2) (Martino et al., 2019) and the Global Ecosystem Dynamics Investigation (GEDI) (Dubayah et al., 2020), both equipped with advanced laser altimetry, demonstrate unprecedented meter or sub-meter accuracy, exhibiting profound potentials on Earth observations at a large scale (Dubayah et al., 2020; Martino et al., 2019). Although ICESat-2 ATL08 data and GEDI L2A data can be combined (Tian and Shan, 2024), the density of the combined measurements is still rather low (less than 100 points/km$^2$) (Lang et al., 2023). The large sparsity of the measurements apparently raises challenges for the subsequential global terrain and forest mapping. Using conventional spatial interpolation methods is only capable of generating a coarse (several hundreds of meters to kilometers of ground spacing) digital elevation model (DEM) over a relatively densely covered region. For instance, ICESat-2 solely is used for DEM generation for Greenland (Fan et al., 2022) and the Antarctic continent (Shen et al., 2022) at 500 m resolution. To derive a finer resolution over regions in low-altitude regions, research integrating ICESat-2 and other Earth observation data has been conducted (Liu et al., 2022; Tian and Shan, 2024, 2023). Therefore, continuous modeling from sparsity data becomes an essential and difficult problem to be solved.

Conventionally, spatial interpolation is a technique to achieve continuous modeling from sparse measurements (Lam, 1983; Mitas and Mitasova, 1999). The most commonly used traditional spatial interpolation methods include Kriging and a series of its variants (Cressie, 1990; Oliver and Webster, 1990; Yin et al., 2022), and the natural neighbor (NN) interpolation, which considers the neighbors of the target location by using Delaunay triangles or Voronoi polygons (Beutel et al., 2010; Musashi et al., 2018; Sibson, 1981). However, the performance of interpolation highly depends on the sparsity of the measurements. As demonstrated by (Buongiorno Nardelli et al., 2016; Chen et al., 2023; Diggle et al., 2002; Urquhart et al., 2013; Wu et al., 2023; Xu et al., 2018; Yao et al., 2024) and many others, the accuracy of spatial interpolation methods may degrade significantly in regions with sparse data. Using the aforementioned spaceborne laser altimetry instances as an example, Fig. 1 presents the DEM interpolation results from measurements with large sparsity using ordinary Kriging (OK) interpolation and NN interpolation methods. To give a better quantitative description on the sparsity of the measurements, we define the sparsity of the discontinuous measurements as the ratio of the number of to-be-interpolated grids (cells) to the number of grids of the DEM. The measurements shown in Fig. 1 include the GEDI L2A and ICESat-2 ATL08 observations over the area (~0.9 km$^2$), and have a 90% sparsity. It demonstrates the fact that OK and NN interpolation methods would exhibit poor performance on large sparsity observations. One of the fundamental reasons for such unsatisfactory behavior is that these conventional methods do not have "memory". Their common assumption is that spatial interpolation beyond a certain neighborhood is mostly independent. We argue this should not be the case, since the similar landscape far away can share the similar topographic pattern. In addition, as more data becomes available, an interpolation method should become smarter, which essentially requires the interpolator to be able to learn from previous operations based on existing data.

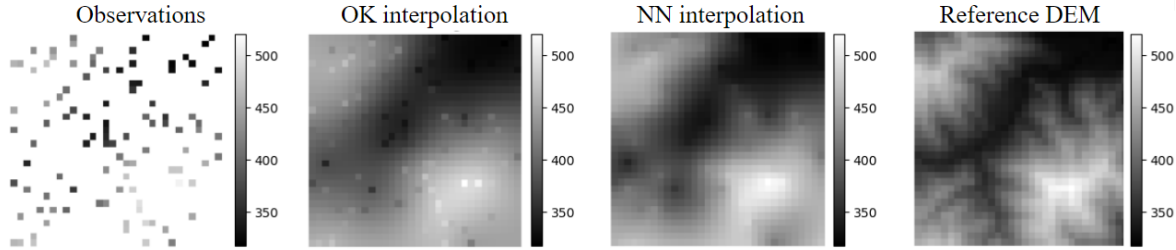

**Fig. 1**. DEM interpolation results from large (90%) sparsity measurements by the ordinary kriging (OK) and natural neighbor (NN) methods, and the reference DEM. The sparsity is the ratio of the number of to-be-interpolated grids to the number of grids of the DEM.

Recent deep learning based methods can compensate for such intrinsic weakness of the conventional interpolation methods (Janowicz et al., 2020; Liu and Biljecki, 2022; Liu et al., 2022; Wu et al., 2023). Such methods have gained substantial interest due to their extraordinary performance, particularly in the context of generative models for image generation tasks (Guérin et al., 2017; Jozdani et al., 2022). Notable achievements have been made on natural images in the computer vision field. The transformation and application of the generative models on handling DEM data is feasible out of two major considerations. Firstly, the grid-based DEM has a similar data structure as the pixel-based natural images (Jozdani et al., 2022; Li et al., 2022; Zhang and Yu, 2022). And secondly, spatial interpolation on sparsely sampled data can be conceptualized as a generative process (Zhang et al., 2022; Zhu et al., 2020). Meanwhile, despite the similarities, it is imperative to acknowledge and address the intrinsic dissimilarities between natural images and DEM data, which manifest in three aspects (Li et al., 2022; Qiu et al., 2019; Zhang et al., 2022): (1) DEM data present complex, 3D characteristics that are unique for topography and other spatial phenomena, e.g., peak, saddle, and valley. For generative models in computer vision, the task for natural images is based on local features (usually the foreground objects). However, topographic details of the entire DEM are the real focus when dealing with the DEM data. (2) Terrain elevation presents a much wider range compared to the pixel intensity values in natural images. Specifically, while RGB values in natural images are each usually in the range of 0-255, terrain elevation values are scale dependent and can span a much wider and arbitrary range. Both the range and distribution patterns of these elevation values vary considerably across different geographical regions, influenced by the diversity of landscape features and geodynamic evolutions. (3) In terms of applications and therefore the requirements, the quality of the DEM needs to be assessed numerically whereas the restored image is mostly for perception or visualization purposes. Any small imperfections in DEM will be recorded by numerical computation, whereas such artifacts or errors in the restored image usually may not affect human interpretation.

Applying generative models to spatial interpolation has been investigated in multiple fields, including DEM generation. Currently, most research integrating generative models and spatial interpolation is based on conditional generative adversarial networks (CGANs) (Li et al., 2022; Mirza and Osindero, 2014; Qiu et al., 2019; Zhang et al., 2022; Zhu et al., 2020), built upon the concept of generative adversarial network (GAN) (Goodfellow and Pouget-Abadie, 2014), a powerful architecture for training image generative models. Many researchers have pointed out that CGANs, utilizing convolution operation, are powerful for extracting underlying patterns given complex spatial contexts (Li et al., 2022; Qiu et al., 2019; Zhang et al., 2022; Zhu et al., 2020). Zhu et al. designed a deep learning model named Conditional Encoder-Decoder Generative Adversarial Neural Networks (CEDGANs) that incorporates an encoder-decoder structure with adversarial learning (Zhu et al., 2020). A case study on terrains demonstrates the model's ability to achieve outstanding spatial interpolation results compared to benchmark methods including inverse distance weighting (IDW) and OK interpolation method. The reconstructed DEM from the CEDGANs exhibits a root mean squared error (RMSE) of 2.5 m compared to 2.9 m from IDW method when the sparsity of the available data reaches 90% of the entire DEM region. Other CGAN based spatial interpolation methods not only require additional topographic knowledge of valleys and ridges besides the observed data (corrupted or incomplete DEM), but also is limited to scenarios where the data sparsity is much smaller (no more than 50%), which means more measurements are needed for interpolation (Li et al., 2022; Qiu et al., 2019; Zhang et al., 2022).

Despite existing research showing that certain CGAN based models exhibit commendable results in spatial interpolation (Li et al., 2022; Qiu et al., 2019; Zhang et al., 2022; Zhu et al., 2020), problems and challenges still remain. (1) A CGAN model, which consists of a generator and a discriminator, can be challenging to train and may suffer from instability issues and are often complex and costly to implement (Zhou et al., 2023). (2) A typical convolutional operation, which is the basis operation used to extract deep features by the above CGAN based model, is proven to only use local features (Brendel and Bethge, 2019). In addition, it also has a relative local/small receptive field, which is an indication of the network's ability to model long range spatial dependencies (Naseer et al., 2021). Thereby, these existing CGAN based models using convolution for feature extraction are not capable of effectively reconstructing terrain features, particularly for data with large sparsity, and more notably have not gained a thorough demonstration on its transferability across different landscapes without additional finetuning.

This paper develops a (vision) Transformer based Generative Model for Spatial Interpolation (T-GMSI) with an objective to optimize the spatial interpolation from highly sparse sampled data. It is inspired from the masking strategy of the masked autoencoder (MAE), which is proposed as a generative model for self-supervised pre-training tactics and shows a great potential to learn deep features from the input images (He et al., 2021). Our proposed model also takes advantage of the strong capability on the long range connection of the vision transformer (Dosovitskiy et al., 2020; Naseer et al., 2021). Our work is based on a customization of the encoder-decoder structure, which is a common network structure for generative models, with the focus on terrain interpolation under high sparsity using the ViT. The ViT is proven to have a larger and global receptive field than convolutional operation (Mao et al., 2021; Naseer et al., 2021). This allows the ViT to model global context and preserve the structural information compared to convolution based networks, and makes the ViT have greater performance towards robustness and generalization of the learned features (Mao et al., 2021; Naseer et al., 2021). Therefore, the evident large receptive fields of the ViT are expected to have great potential on handling interpolation on high sparsity input data.

The new development of the proposed model includes an asymmetric encoder-decoder structure using a masking strategy, a ViT based feature extraction module, and a new loss module, which combined intends to gain a robust transferability. Specifically, the contributions of our study can be summarized as follows:
  (1) Instead of convolution operation used in typical generative model, we construct a ViT-Base (Dosovitskiy et al., 2020) based encoder to extract features of the input data. This replacement makes our model inherit the larger receptive field of the ViT (Naseer et al., 2021) and be able to learn the long range connection between the sparsely distributed input data, which ensures the robustness and transferability across different geographical locations of the proposed model.

(2) We replace the patch embedding, which is a necessary and critical procedure in the ViT for natural images, with grid embedding for the ViT implemented in our encoder. This customization makes the model suitable for spatial interpolation considering the grid-based DEM as input training data.
(3) Instead of using the reconstruction loss functions solely for natural image generative models, we develop a terrain feature aware loss function. This loss function assures a better estimation of the actual topography rather than a simply realistic-looking reproduction.
(4) We produce and publicly share a well-pretrained T-GMSI model specializing in DEM generation. This model is proven to have well transferability across diverse landscapes and is capable of handling measurements with a range of sparsity levels.

The performance of the proposed T-GMSI is tested with interpolating digital terrain data for space altimeter GEDI and ICESat-2 measurements, where the footprint point spacing is often at the magnitude of hundreds to thousands meters (Dubayah et al., 2020; Neumann et al., 2019). Our goal is to use these measurements and a pre-trained model to produce a DEM with a resolution of at least tens of times finer. We demonstrate that the proposed model is able to generate continuous fine elevation surface from large sparsity data (sparsity > 70%) and achieve state-of-the-art (SOTA) performance. Most notably, its superior transferability is verified by applying this approach to multiple different types of large-scale terrains. The developed model can be regarded as a new paradigm for spatial interpolation and adopted to other large sparsity data interpolation tasks beyond DEM generation.

The remainder of this paper is organized as follows. Section 2 describes our developed method in detail and the experimental evaluation methods. Section 3 presents the results and comparison analysis against several baseline methods. Section 4 discusses the accuracy of the proposed method under different experimental setups and the transferability of the pretrained model to diverse landscapes. Section 5 presents findings and proposes future work. The source code we have developed to implement our method, the pre-trained model as well as the accomplished products of the generated DEM is shared on https://github.com/XXXXX (to release upon acceptance).

## 2. Materials and methods

By customizing the masked autoencoder and the ViT-Base (Dosovitskiy et al., 2020; He et al., 2021), we develop a generative model for spatial interpolation using transformer for continuous DEM mapping from highly sparse measurements. In this section, we will describe the architecture of our T-GMSI in detail and the experimental setup for its implementation.

**2.1 The T-GMSI model**

As shown in Fig. 2, the proposed T-GMSI is a generative model based on an asymmetric encoder-decoder structure for spatial interpolation using ViT architecture. The core of the architecture is built upon the ViT framework, which has been adapted to handle the unique characteristics of grid-based DEM data. The model consists of a transformer encoder and a transformer decoder, both leveraging the capabilities of the ViT-Base configuration. The input to the model is a DEM tile, which undergoes a grid embedding process to transform the DEM tile into a suitable format for the transformer encoder. The transformer encoder comprises multiple transformer blocks, each containing a layer normalization, multi-head attention mechanism, and a multi-layer perceptron (MLP). This series of transformer blocks allows the model to learn complex spatial relationships within the DEM tile.

Fig. 2 also illustrates the training and inferring process. To train the model, we apply the masking strategy (He et al., 2021; Xie et al., 2021) to intrinsically simulate the sparsity of the sparse measurements for interpolation and force the model to learn the relationship between the long range separated grids. The model is provided with multiple DEM tiles and each DEM tile (size: 32✕32 grids) is first grid embedded and positional embedded. At each training epoch, a random mask at a certain mask ratio is applied on the embedded DEM tile, obtaining a masked DEM tile. The mask ratio is defined as the ratio of the masked/removed grids to the total number of grids of an input DEM tile. To clarify, the term of 'mask ratio' is only referred when applying the masking strategy during training procedure, while the term 'sparsity or sparsity level' is always referred to the low density of the measurements to-be-interpolated. After masking, the masked and embedded DEM tile is fed into the encoder and decoder. Since the dimension of the intermediate input to the encoder and the decoder is different, the encoder-decoder structure in our method is asymmetric. After the training procedure, the pre-trained model can be used for inferring, i.e., interpolation, in which an unseen corrupted DEM tile (size: 32✕32 grids) at arbitrary sparsity level can be interpolated by feeding it into the pre-trained model. The inferring procedure of the unseen corrupted DEM is demonstrated in Fig. 2 as well. The following paragraphs will demonstrate these processes in detail along with the training procedure.

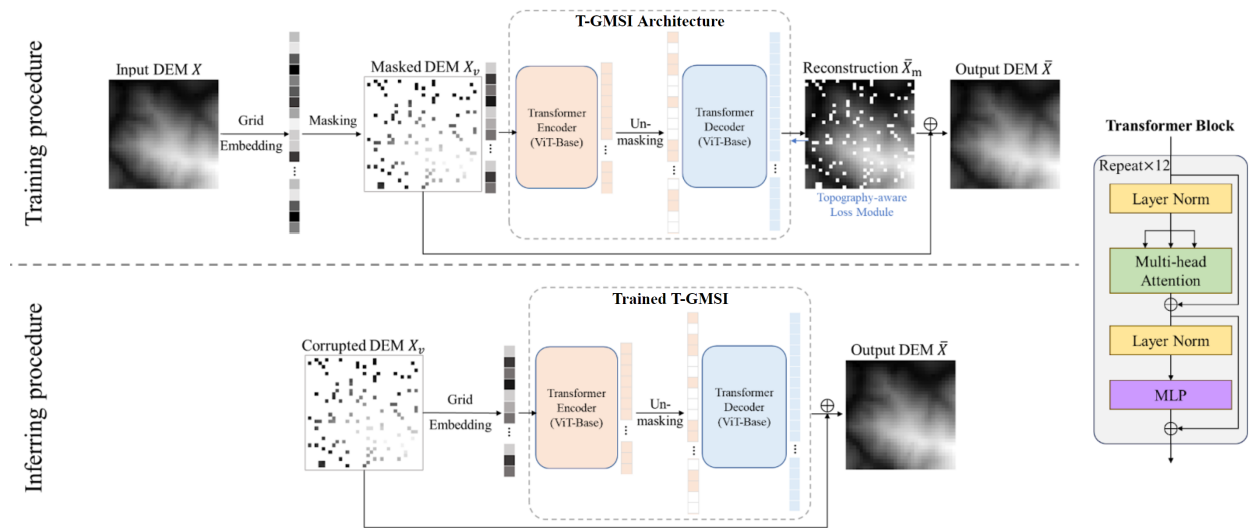

**Fig. 2**. The architecture of the T-GMSI, and its training and inferring procedure. The training procedure follows five steps, consisting of 1) grid and positional embedding, 2) masking, 3) transformer encoding, 4) unmasking and transformer decoding, and 5) loss calculation. To conduct inferring on a to-be-interpolated DEM for interpolation, the procedure follows three steps: 1) grid and positional embedding, 2) transformer encoding, and 3) unmasking and transformer decoding.

**Grid embedding and positional embedding.** Similar to a standard ViT that divides an image into patches (patch embedding) with added positional embeddings to each patch, we flatten the input DEM tile by grids which is referred to as grid embedding in this paper and with added positional embeddings to each grid. A standard ViT cannot take the raw image as input but uses patch embedding. It divides a natural image into multiple patches, each of which consists of 16✕16 pixels, and uses an sinusoidal to positionally embed each patch (Dosovitskiy et al., 2020; Vaswani et al., 2017). Keeping the dissimilarities between the natural images and grid-based DEM in mind, we use grid embedding for the grid-based DEM tile instead of patch to embed the DEM tile before it is masked and fed into the transformer encoder. To be specific, grid embedding is equivalent to patch embedding with patch size 1, which means the 32×32 grids DEM tile is transformed into a 1×1024 flatten vector grid by grid as shown in Fig. 2 above; each grid then undergoes sinusoidal positional embedding (Vaswani et al., 2017).

**Masking strategy for training.** After the input DEM is grid embedded and positional embedded, a masking strategy (He et al., 2021; Xie et al., 2021) is used to mask/corrupt the DEM to obtain the masked DEM (i.e., the masked DEM shown in Fig. 2). Since random masking is proven to be the best masking strategy for self-supervised learning in several studies (He et al., 2021; Xie et al., 2021), it is chosen in our study as well. The random masking strategy is a commonly used data sampling strategy and straightforward. It randomly masks/removes some grids (at a specific mask ratio, i.e., the ratio of the masked grids to the total number of grids of the input DEM) without replacement (i.e., one patch will only be selected once) and keeps the remaining grids as the input to the subsequent encoder. For a larger mask ratio, fewer available grids are fed into the encoder, which means less detailed spatial information is given for continuous elevation mapping. In addition, to explore how different mask ratios used during training would impact the performance of the trained interpolator, a series of mask ratios will be used to train the model whose performance will evaluated through assessing the resultant DEMs.

**Encoder.** The encoder of the proposed model is ViT-Base based, which consists of 12 transformer blocks with 768 embedding dimensions (Dosovitskiy et al., 2020). The encoder is only applied on the visible/unmasked grids in the masked DEM, which maps the masked DEM to a latent representation. As the result, the input of our T-GMSI encoder is all visible grids with grid embedding and positional embedding. The advantages of our encoder are two folded. 1) The conduction of the ViT makes our model have a larger receptive field than other convolution based interpolation models, and 2) the grid embedding procedure leverages all visible grids in the masked DEM efficiently.

**Decoder.** As for the decoder of T-GMSI that reconstructs the DEM from the latent representation, it is also ViT-Base based, consisting of 12 transformer blocks with 768 embedding dimensions (Dosovitskiy et al., 2020). Compared to the original mask autoencoder paradigm (8 transformer blocks and 512 embedding dimension) (He et al., 2021), our decoder is much deeper and more complex. Different from the encoder, the decoder is applied on all grids (visible and invisible). Specifically, the grid embeddings and positional embeddings of the invisible grids are inserted back into the output of the encoder, which is denoted as the unmasking process in Fig. 2 and fed into the decoder. The decoder intends to precisely reconstruct the masked/corrupted DEM and preserve the topographic details.

**Topography-aware loss module.** In the training procedure of the proposed T-GMSI, a topography-aware loss module is conducted for a better reconstruction for the elevation mapping from the perspective of terrain feature optimization. Typical generative models designed for natural images use mean squared error (MSE) loss to monitor and steer the training. However, it has been reported that it may not be able to comprehensively address the unique characteristics of the topography and might output overly smoothed images that lack high-frequency topographic details (Ledig et al., 2017; Zhang et al., 2019, 2022). Therefore, we intend to introduce the gradient (i.e., terrain slope) loss to guide the training procedure of the model towards the DEM reconstruction. There are two parts in our topography-aware loss modules: the MSE loss which aims to control the global terrain accuracy, and the gradient loss which is designed to preserve local terrain characteristics. Combining the MSE loss and gradient loss, the proposed topography-aware loss module is determined by their weighted summation:

$$\mathcal{L}_{reconstruction} = \mathcal{L}_{MSE} + \gamma \mathcal{L}_{gradient} \quad (1)$$

where $\gamma$ denotes the weight of the gradient loss. The first term in Equation (1) is the MSE loss, denoted as $\mathcal{L}_{MSE}$, which is adopted to optimize the global terrain accuracy:

$$\mathcal{L}_{MSE} = \frac{1}{W \times H} \sum_{i=1}^{W} \sum_{j=1}^{H} (\hat{h}_{i,j} - h_{i,j})^2 \quad (2)$$

in which $W$ and $H$ are the width and height of the input DEM tile, $\hat{h}$ is the known terrain height from the input DEM, and $h$ is the reconstructed terrain height from the decoder. The second term in Equation (1) is the gradient loss $\mathcal{L}_{gradient}$, which represents the smoothness of the terrain. For the reconstructed DEM, the gradient at grid $i, j$, denoted as $s_{i,j}$ can be calculated in a differentiable way as follows:

$$s_{i,j} = arctan \sqrt{dx_{i,j}^2 + dy_{i,j}^2} \quad (3)$$

in which the slope $dx_{i,j}$ and $dy_{i,j}$ are calculated with Sobel operators. By comparing the gradient of the reconstructed DEM and the gradient of the input DEM, the gradient loss $\mathcal{L}_{gradient}$ can be obtained by

$$\mathcal{L}_{gradient} = \frac{1}{W \times H} \sum_{i=1}^{W} \sum_{j=1}^{H} (\hat{s}_{i,j} - s_{i,j})^2 \quad (4)$$

where $\hat{s}$ is the gradient of the input DEM tile, and $s$ is the calculated gradient of the reconstructed DEM from the decoder.

## 2.2 Evaluation method

To evaluate the performance of the proposed interpolator, its results are compared with the results interpolated by OK and NN interpolation methods, which are considered as the baseline methods for spatial interpolation. In addition, the open source CGAN-based method CDEGAN proposed by Zhu et al. (Zhu et al., 2020) will be conducted for comparison as well. The evaluation is performed with reference to the reference DEM introduced in the following section. The comparison between different methods is conducted from visual examination as well as quantity investigation. Visual examination is applied to establish the overall achievement of the interpolated DEM. The quantity investigation at grid level is accomplished from three aspects. (1) elevation assessment, (2) slope assessment, and (3) stream assessment; all between the resultant DEM and the reference DEM. The resultant DEM includes the results from the proposed model and the baseline methods.

Unlike many popular studies, we also assess the quality of the derived streams from the resultant DEMs. To this end, the reference stream network is extracted from the reference DEM following a two-step procedure. (1) Extract stream grids from the reference DEM using a specified accumulation threshold, and (2) buffer the extracted stream grids by one grid as the final reference stream network. When performing the stream assessment on the resultant DEMs, the stream grids are extracted from the interpolated DEMs using the identical accumulation threshold as the one used to obtain the reference stream network. Then, the stream grids from interpolated DEMs falling into the reference stream network will be denoted as the true positive stream grids, while the stream grids outside the reference stream network will be denoted as the false positive stream grids. In such a way, the true positive (TP) stream, false positive (FP) stream, and false negative (FN) stream can be counted. Then, the stream assessment will be conducted using the two common metrics, precision (TP/(TP+FP)) and recall (TP/(TP+FN)). In addition, to assure a comprehensive evaluation we also assess the streams at different scales using different accumulation thresholds.

## 2.3 Study areas and datasets

As shown in Fig. 3, four entire counties in the United States are selected as the study areas, including the Mendocino County in California State (CA), Piute County in Utah State (UT), Sanborn County in South Dakota State (SD), and Tippecanoe County in Indiana State (IN). These counties consist of various landscapes that have a distinguished range of terrain height. The first two counties (Mendocino and Piute) are representatives of mountainous regions with rugged terrain, while the last two counties (Sanborn and Tippecanoe) are examples of plain regions with relatively flat topography.

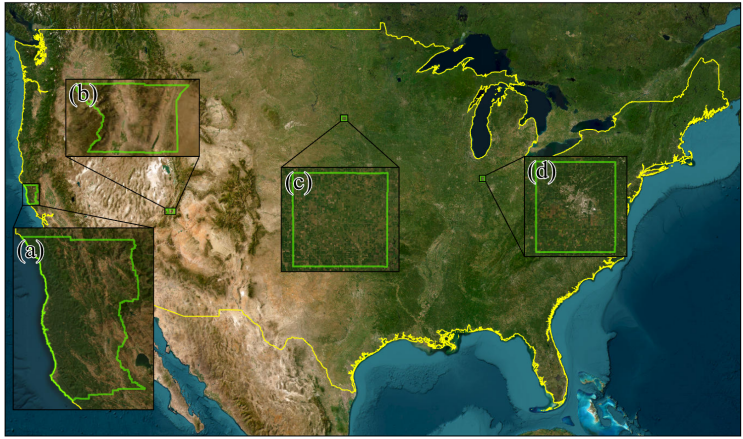

**Fig. 3**. Study areas in the US for experiments on spatial interpolation, including (a) Mendocino County, CA, (b) Piute County, UT, (c) Sanborn County, SD, and (d) Tippecanoe County, IN. The (a) and (b) areas are mountainous while (c) and (d) are plain. The total size of the four counties is 14,634 km$^2$.

**Table 1**. Summary of the study areas

|  |  | Mendocino, CA | Piute, UT | Sanborn, SD | Tippecanoe, IN |
|---|---|---|---|---|---|
| Area[1] (km$^2$) | | 9,852 | 1,980 | 1,500 | 1,302 |
| Average and standard deviation of elevation[2] (m) | | 579±395 | 2303±360 | 396±6 | 207±21 |
| Maximum elevation difference[2] (m) | | 2100 | 1808 | 51 | 110 |
| GEDI L2A v002 | Number of data points | 701,664 | 154,077 | 95,658 | 61,956 |
| | Elevation accuracy[3] (m) | -0.47±15.21 | -0.38±7.03 | -0.26±0.70 | -0.21±1.88 |
| ICESat-2 ATL08 v006 | Number of data points | 125,174 | 34,551 | 60,368 | 31,484 |
| | Elevation accuracy[3] (m) | -1.38±5.56 | -1.23±3.42 | -0.68±0.38 | -0.53±0.96 |
| Data density[4] (points/km$^2$) | | 83.93 | 95.27 | 104.02 | 71.77 |

[1] The areas of the counties are collected from *Census 2020*.
[2] The statistics of the elevation are summarized from the 1 m 3DEP (3D Elevation Program) DEM.
[3] The elevation accuracy is obtained by comparison with the 1 m 3DEP DEM.
[4] Data density is obtained from the combination of GEDI and ICESat-2.

Table 1 summarizes the characteristics of these four study areas. In our study, Mendocino County is used to construct the training dataset for the model training and obtain the pre-trained model. The other three study areas are selected to explore the transferability of the pre-trained model. Piute County shares a similar rugged topography with Mendocino County where the training data locates, while the other two (Sanborn and Tippecanoe County) present a rather distinguished smooth and plain landscape. The assessment on the resultant DEM using the pre-trained model over these study areas will be conducted with reference to a high accuracy reference DEM.

*2.3.1 Data for model training*

The fine resolution DEM of USGS 3D elevation program (3DEP) (Arundel et al., 2015) over Mendocino County is used to construct the training dataset to test the feasibility and effectiveness of the proposed method. These 3DEP DEMs at 1 meter resolution are produced from airborne lidar under the USGS National Elevation Program for the bare ground elevation and distributed for easy public access through the Google Earth Engine (GEE) Data Catalog (Arundel et al., 2015; Gorelick et al., 2017). According to the specification of the 3DEP provided by USGS, its vertical accuracy is 10 cm (1σ) in the non-vegetated areas and 15 cm (1σ) in the vegetated areas (Arundel et al., 2015). The 1 m 3DEP DEM is collected from GEE and down-sampled to 30 m resolution to form the training dataset and to evaluate the interpolated 30 m DEMs using different interpolation methods. After the projection of coordinate reference system to the corresponding state plane coordinate system, the entire 30 m 3DEP DEM in Mendocino County, which covers 9,852 km$^2$ area as shown in Table 1, are cropped into multiple single-channel DEM tiles (with each being 1×32×32 grids) with 50% frontal and 50% side overlap, which means no overlap, as the DEM tiles for model training. Around 370,000 DEM tiles are obtained for Mendocino County, with 330,000 tiles (89%) for training set and 40,000 tiles (11%) for validation set.

*2.3.2 Data for model prediction*

The input data for model prediction (i.e., interpolation) is sparsely distributed elevation measurements. As a realistic application, we use the combined footprints of GEID and ICESat-2 as a representation of such sparse distribution. Two datasets are prepared. For one dataset, the elevations at these footprints are directly from the elevation measurements of space borne lidar GEDI and ICESat-2, while for the other dataset, its elevations are from 3DEP DEM collected by airborne lidar.

**1) Spaceborne lidar measurements**. The combination of GEDI and ICESat-2 data, including their footprints and heights, forms the set of spaceborne lidar measurements with large sparsity. Since there exists measurement errors (terrain height differences) in the GEDI and ICESat-2 data comparing to the reference 3DEP DEM, assessments on the resultant DEMs generated from space borne lidar as real measurements using different models will be used to investigate the capability of the proposed method to perform spatial interpolation on erroneous large sparsity data from space.

The GEDI L2A data with the lowest return height (*elev_lowestmode*) acquired from Google Earth Engine Data Catalog (Gorelick et al., 2017) is regarded as the terrain height (transformed from ellipsoid height to orthometric height). The footprints of GEDI L2A are spaced ~600 m cross track and ~60 m along track with a ~25 m diameter (Dubayah et al., 2020). The data were collected from March 25 of 2019 to June 1 of 2022, and a filtering process following suggested criteria in previous research (Dubayah et al., 2020; Liu et al., 2021; Shendryk, 2022; Tian and Shan, 2024) is conducted to assure the quality of the input measurements.

Beside the GEDI L2A data, we also collect the ICESat-2 ATL08 terrain height information (*h_te_bestfit*) downloaded from the National Snow and Ice Data Center (NSIDC) through the Python package *icepyx* (The icepyx Developers, 2023), which provides users with valuable terrain height information with a global coverage at a fixed segment size of 100 m (Tian and Shan, 2021). The data were collected from October 2018 to September 2022 for all four study areas. A two-step filtering procedure is also performed. In the first step, the ATL08 data with an invalid value (3.4028E+38) of *h_te_uncertainty* are first excluded (Neuenschwander and Pitts, 2019; Tian and Shan, 2021). In addition to the invalid value, it is shown from the previous study that the larger the uncertainty index, the less reliable the terrain height of the specific points (Tian and Shan, 2021). Therefore, points with *h_te_uncertainty* larger than the 95% percentile of the total distribution over the study areas are excluded as well.

The elevation accuracy of the spaceborne lidar measurements from GEDI and ICESat-2 after the above preprocessing are summarized in Table 1.

**2) Aerial lidar measurements**. As listed in Table 1, there are obvious errors or difference for the spaceborne lidar measurements of the terrain height from ICESat-2 and GEDI comparing to the reference 3DEP DEM. Therefore, by preparing the lidar measurements, we aim to eliminate the effect of the offsets on the terrain height measurements of the ICESat-2 and GEDI meanwhile retaining their spatial pattern. Combining the terrain height of the 30 m 3DEP DEM created from airborne lidar, and geolocation of spaceborne lidar footprints from GEDI and ICESat-2, we generate a set of ideal lidar measurements with large sparsity. To be specific, the ideal lidar measurements are composed of the terrain heights extracted from the 30 m 3DEP DEM at the exact geolocation of the space lidar footprints. Utilizing the airborne lidar as ideal measurements, we aim to examine the best possible performance of the proposed model on spatial interpolation of large sparsity data. When airborne measurements are used, the interpolation results are expected to reflect the intrinsic performance of the model since it is trained by a set of 30 m 3DEP DEMs that have the same quality as these measurements.

## 3. Results

Through the training using 95% mask ratio with the 30 m DEM of Mendocino County, the trained T-GMSI95 is obtained by examining the loss on the training dataset and validation dataset. In this section, multiple 30 m DEMs are generated respectively from the airborne lidar measurements and the spaceborne lidar measurements in Mendocino County by using OK interpolation method, NN interpolation method, and the trained T-GMSI95, i.e., the T-GMSI trained at 95% mask ratio. Comparison on the performance of these methods is conducted by a comprehensive assessment of these DEMs using the evaluation metrics described above.

### 3.1 Elevation assessment

To clearly show the quality of the generated DEMs from different methods, three sub areas within Mendocino County are selected for detailed visual comparison on the DEMs. As shown in Fig. 4 (a), the OK method exhibits unsatisfying performances on large sparsity lidar measurements for both airborne and spaceborne lidar measurements, and the interpolation working principle leads to an even poorer results dealing with the spaceborne lidar measurements with significant noises, e.g., the Area 3 shown in Fig. 4 (a). For the NN interpolation method, though it has a better visual effect than the OK method, it over smooths the landscape and lacks a great deal of detail compared to our model.

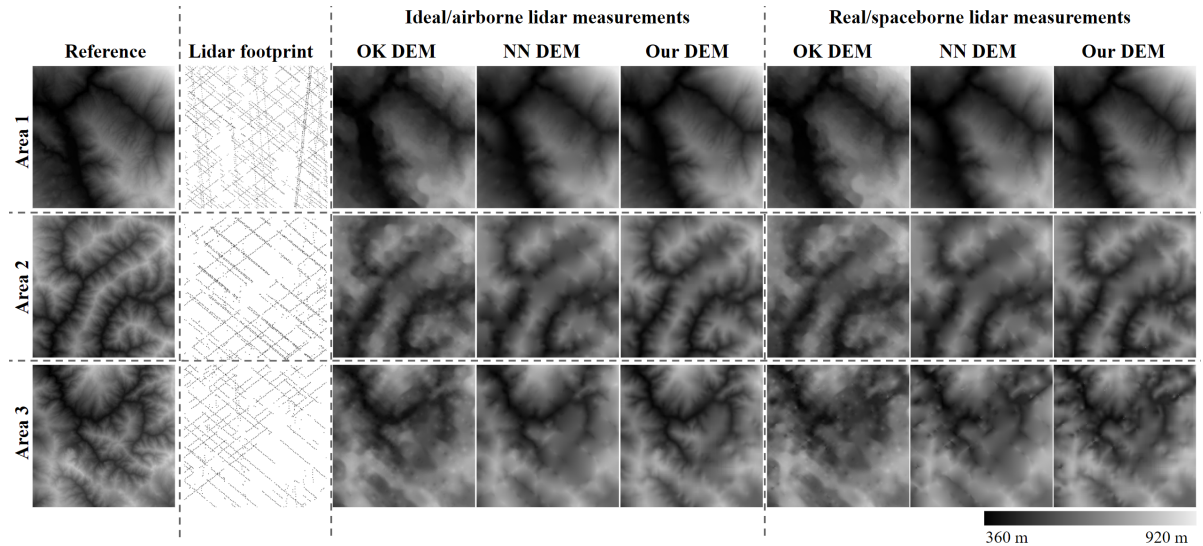

(a) Elevation assessment.

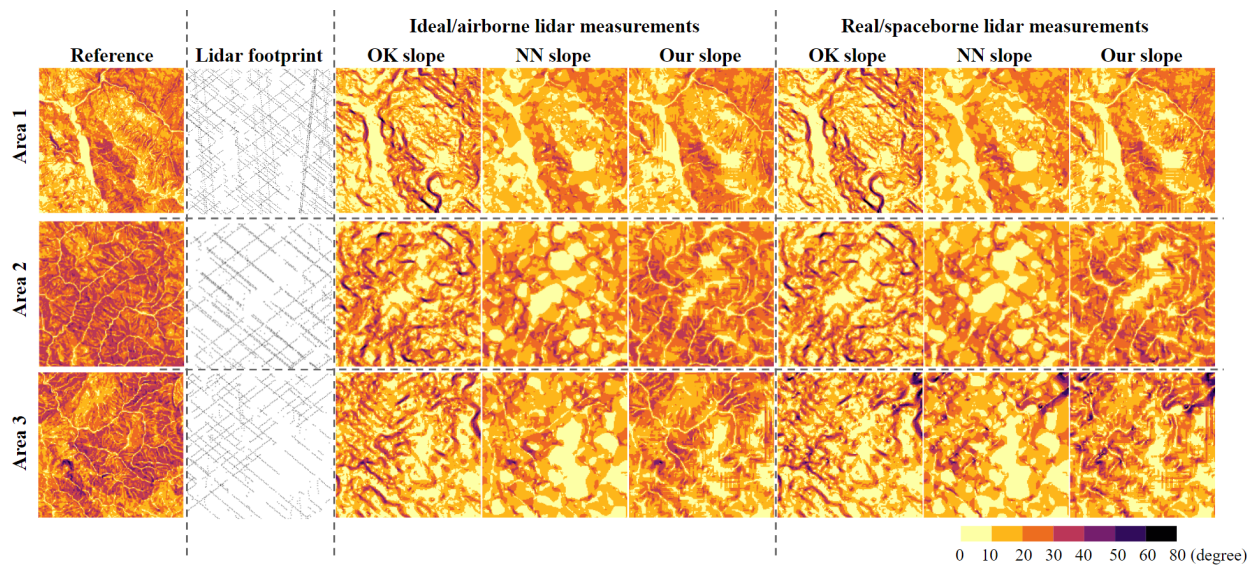

(b) Slope assessment.

**Fig. 4**. Elevation (a) and slope (b) assessment of the resultant DEMs over three selected sub areas in Mendocino County from different interpolation methods. From left to right: the reference 30 m 3DEP DEM, distribution of the lidar footprints as sample locations, and the elevation (a) or slope (b) from the interpolated DEMs using airborne lidar measurements and spaceborne lidar measurements with OK, NN and our methods. The sizes of these sub areas are ~30 km$^2$ each, and the sparsity level of the samples are ~90%, 94%, and 96% for the three sub areas, respectively.

Table 2 is a summarized comparison between different methods by providing the assessment statistics of the resultant DEMs over the entire Mendocino County. The best results from the two sets of lidar measurements are bolded in the table. For the airborne lidar measurements, the proposed method achieves a significant improvement compared to the OK and NN interpolation methods, with a drop in root mean square error (RMSE) of 40% and 25%, respectively. When the input measurements contain noises (real spaceborne lidar measurements) comparing to the training dataset, though subtler than the airborne lidar measurements, the superiority of the proposed method still exists with a 23% and 10% decrease on the RMSE compared to the conventional OK and NN interpolation method. By experimenting on the airborne lidar measurements, the best performance of the proposed model can be acknowledged. As shown in Table 2, on the one hand, the improvement achieved by our model on airborne measurements gives the best quality the model can obtain, which shows a large superiority. This is governed by the intrinsic capability of the proposed model. On the other hand, the results on spaceborne lidar measurements proves the capability of the proposed model handling outliers during the interpolation procedure, which is still rather remarkably better than the baseline methods.

**Table 2**. Elevation and slope assessment of the resultant 30 m DEMs from conventional interpolation methods and the proposed model in Mendocino County where the model is trained. Bolded numbers indicate the best results.

|  |  | Ideal/airborne lidar measurements | | | Real/spaceborne lidar measurements | | |
| --- | --- | --- | --- | --- | --- | --- | --- |
|  |  | OK | NN | T-GMSI95 | OK | NN | T-GMSI95 |
| Elevation difference (m) | $Mean_{\Delta h}$ | -2.15 | -1.62 | **-0.84** | -2.84 | -2.18 | **-1.66** |
|  | $STD_{\Delta h}$ | 26.26 | 20.91 | **15.91** | 29.86 | 25.53 | **23.11** |
|  | $MAE_{\Delta h}$ | 17.02 | 13.27 | **10.13** | 19.43 | 16.03 | **14.2** |
|  | $RMSE_{\Delta h}$ | 26.35 | 20.97 | **15.93** | 29.99 | 25.63 | **23.17** |
| Slope difference (°) | $MEAN_{\Delta s}$ | -4.83 | -5.16 | **-2.79** | -4.9 | -5.18 | **-3.14** |
|  | $STD_{\Delta s}$ | 9.37 | 7.56 | **6.28** | 9.78 | 8.24 | **7.69** |
|  | $MAE_{\Delta s}$ | 8.03 | 6.74 | **4.92** | 8.36 | 7.19 | **6.05** |
|  | $RMSE_{\Delta s}$ | 10.54 | 9.15 | **6.87** | 10.94 | 9.73 | **8.31** |

### 3.2 Slope assessment

Similar to the assessment for the elevation, Fig. 4 (b) shows the reference slopes derived from the reference 30 m 3DEP DEM and derived slopes from the generated DEMs using different methods over the same three sub areas. Compared to the baseline methods (OK and NN interpolation methods), the slopes from the developed T-GMSI demonstrate the most satisfying performance with very similar visual presentation with the reference slope. The slopes from the proposed method are the ones least affected by the noises in the spaceborne lidar measurements. The OK interpolation method apparently over smooths the topography, while the NN interpolation method under smooths the places where terrain height varies, especially for example of the sub area 2 and 3 in Fig. 4 (b). The same conclusion can be made from Table 2, which summarizes the statistics of the slope difference of the resultant DEMs over the entire Mendocino County. In the comparison of the slope difference, the proposed method demonstrates considerable enhancement over baseline methods in airborne lidar measurements, reducing the RMSE by 35% and 25% compared to the OK and NN interpolation methods, respectively. Furthermore, with noise-prone spaceborne lidar measurements, the method maintains its superiority, exhibiting a 25% and 15% reduction in RMSE of slope difference relative to the OK and NN methods. Finally, it should be noted that these statistical improvements in slope assessment are consistent with the ones in elevation assessment.

**3.3 Stream assessment**

Fig. 5 illustrates the evaluation of streams derived from different DEMs, where the accumulation threshold is set at 2,000,000 m². Detailed visual examinations of the stream difference are conducted across the same three sub areas. Compared to the baseline methods, the T-GMSI demonstrates superior alignment with the reference streams, evident in two key aspects. First, it exhibits the fewest false positives (erroneously identified stream grids) and false negatives (missed stream grids) in its resultant DEMs. Secondly, it has the highest number of true positives (correctly identified stream grids). This conclusion is further supported by Table 3, which summarizes the *precision* and *recall* metrics for the stream evaluation of the resultant DEMs. The table highlights the best results, which are all from the proposed method. In terms of stream evaluation metrics, the proposed T-GMSI consistently outperforms the other two baseline methods, both for airborne, and spaceborne, noise-affected lidar measurements.

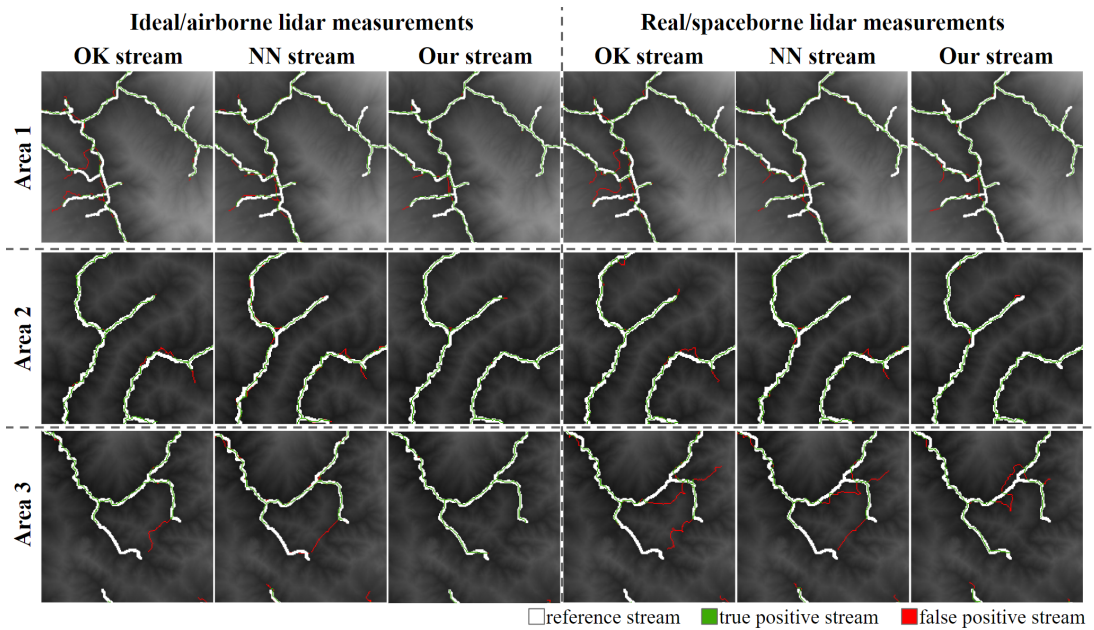

**Fig. 5.** Illustration of the stream assessment over three sub areas (from top to bottom) in Mendocino County created from airborne lidar measurements and spaceborne lidar measurements using different methods (from left to right). The accumulation threshold for stream extraction is 2,000,000 m².

**Table 3.** Stream assessment of the resultant 30 m DEMs from conventional interpolation methods and the proposed model in Mendocino County where the model is trained. The total number of stream grids in the reference DEM is 153,115 and the accumulation threshold for stream extraction is 2,000,000 m². The bolded numbers indicate the best results.

|  | Ideal/airborne lidar measurements | | | Real/spaceborne lidar measurements | | |
| --- | --- | --- | --- | --- | --- | --- |
|  | OK | NN | T-GMSI95 | OK | NN | T-GMSI95 |
| TP | 85,260 | 89,221 | 102,413 | 83,760 | 88,560 | 96,397 |
| TP+FP | 131,629 | 122,521 | 123,803 | 139,557 | 130,788 | 126,118 |
| *precision* | 0.678 | 0.728 | **0.827** | 0.6 | 0.677 | **0.764** |
| *recall* | 0.583 | 0.583 | **0.669** | 0.547 | 0.578 | **0.63** |

Using different accumulation thresholds, distinguished streams can be derived from a DEM. Therefore, we perform the same stream difference evaluation process using two more accumulation thresholds (20,000 m² and 100,000 m² in addition to the 2,000,000 m² above) and calculate the precision and recall accordingly. Fig. 6 presents a plot comparing the precision and recall of three interpolation methods for stream extraction over varying accumulation thresholds. The x-axis represents the accumulation threshold for stream extraction, while the y-axis represents the precision and recall values, ranging from 0.3 to 0.9. From this figure, though the precision and recall vary as the accumulation threshold increases, the proposed model appears to have a higher precision and recall for the entire threshold range compared to the other baseline methods. Overall, the proposed T-GMSI outperforms the OK and NN methods in terms of precision and recall by at least 10% and demonstrates most stable quality across varying accumulation thresholds.
.

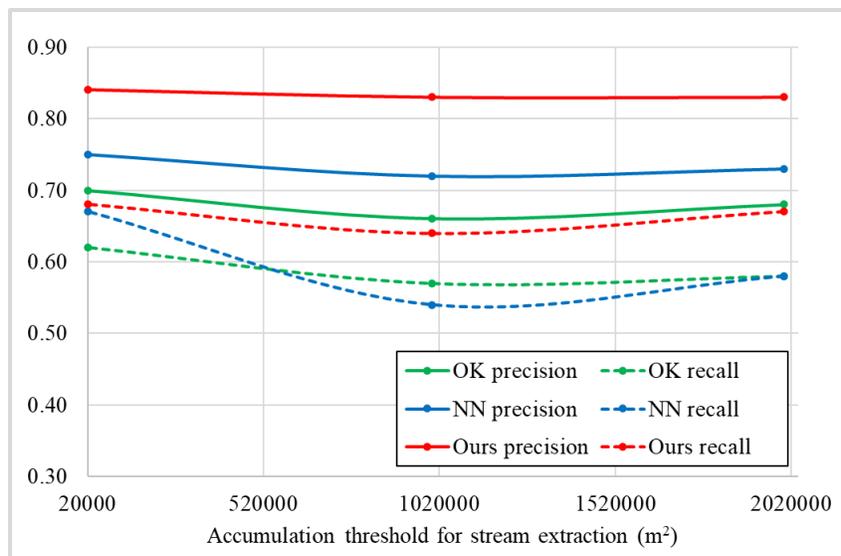

**Fig. 6**. Variation of the precision and recall in stream assessment under varying accumulation thresholds for DEMs from different interpolation methods.

## 4. Discussion

This section presents a comprehensive discussion on the properties of the proposed model. At first, to reveal the effect of different mask ratio during training, a series of mask ratios will be used to train multiple proposed models. Then, the transferability of the pre-trained model with the Mendocino County will be experimented on the three other counties with distinct topography than the training county. At last, our pre-trained model will be compared with an existing pre-trained CGAN-based interpolation model.

### 4.1 Effect of mask ratio for training

So far, the comparison between the interpolated DEMs from different methods is referring to the proposed model pre-trained by using random masking at 95% mask ratio, the T-GMSI95. However, the value of the mask ratio can be regarded as a hyperparameter as an external configuration variable in machine learning. Therefore, in this section, several versions of the pre-trained model using random masking at a different mask ratio ranging from 50% to 98% are evaluated. The aim is to find out the most appropriate mask ratio for training the proposed model.

Table 4 below presents the elevation assessment results of the resultant 30 m DEMs in Mendocino County on airborne lidar measurements using the proposed model pre-trained by different mask ratios. Though there is a mix of slight positive and negative means and different mask ratios lead to different biases in the elevation interpolation, the model trained at 60%, 70%, 80%, and 98% mask ratios give the most unsatisfying results. The column of mean absolute error ranges from 10.13 m to 18.41 m. Except for 98% mask ratio, models trained with higher mask ratio tend to have lower mean absolute error values, indicating a better performance with increased mask ratio during the training procedure and it reaches the minimal at a mask ratio 95%. For the T-GMSI98, its performance decreases sharply. For the standard deviation of the elevation difference, the values range from 15.91 to 31.70 m, with the model at higher mask ratio generally associated with lower standard deviation. Similar to the mean absolute error and standard deviation, the RMSE ranges from 15.93 m to 31.93 m, with lower RMSE generally seen for models trained with higher mask ratio.

**Table 4**. Summary of the elevation assessment on the resultant 30 m DEMs of Mendocino County interpolated with airborne lidar measurements by the proposed model trained with different mask ratios.

| Quality metrics | Trained T-GMSI model using different mask ratios | | | | | | |
|---|---|---|---|---|---|---|---|
| | T-GMSI50 (50%) | T-GMSI60 (60%) | T-GMSI70 (70%) | T-GMSI80 (80%) | T-GMSI90 (90%) | T-GMSI95 (95%) | T-GMSI98 (98%) |
| $Mean_{\Delta h}$ (m) | 0.95 | -3.83 | -6.23 | -7.16 | **-0.19** | -0.84 | -4.62 |
| $STD_{\Delta h}$ (m) | 17.98 | 31.70 | 17.85 | 18.08 | 18.05 | **15.91** | 27.99 |
| $MAE_{\Delta h}$ (m) | 10.24 | 18.41 | 11.71 | 12.27 | 10.92 | **10.13** | 13.16 |
| $RMSE_{\Delta h}$ (m) | 18.00 | 31.93 | 18.91 | 19.45 | 18.05 | **15.93** | 28.37 |

These metrics across different mask ratios (50% to 95%) suggest that training with higher mask ratio generally leads to lower error and variability in elevation interpolation. Visually, Fig. 7 shows the three sub areas of the interpolated DEMs from the pre-trained model using different mask ratios. In general, as the mask ratio during training increases, the model can be better trained and its resultant DEM becomes more detailed and less abstractive, with the model trained at 95% mask ratio showing the best results.

In summary, the table quantitatively and the figure qualitatively demonstrate the impact of selecting mask ratio at the training stage on the accuracy and visual quality of interpolated DEMs. Choosing a higher mask ratio for training often associates with decreased error metrics (mean absolute error and RMSE) and variability (STD), suggesting that models trained with larger mask ratio may offer a better balance between detail and accuracy in elevation interpolation. The figure visually illustrates these changes, with increasing mask ratio during training leading to more defined and less noisy DEMs.

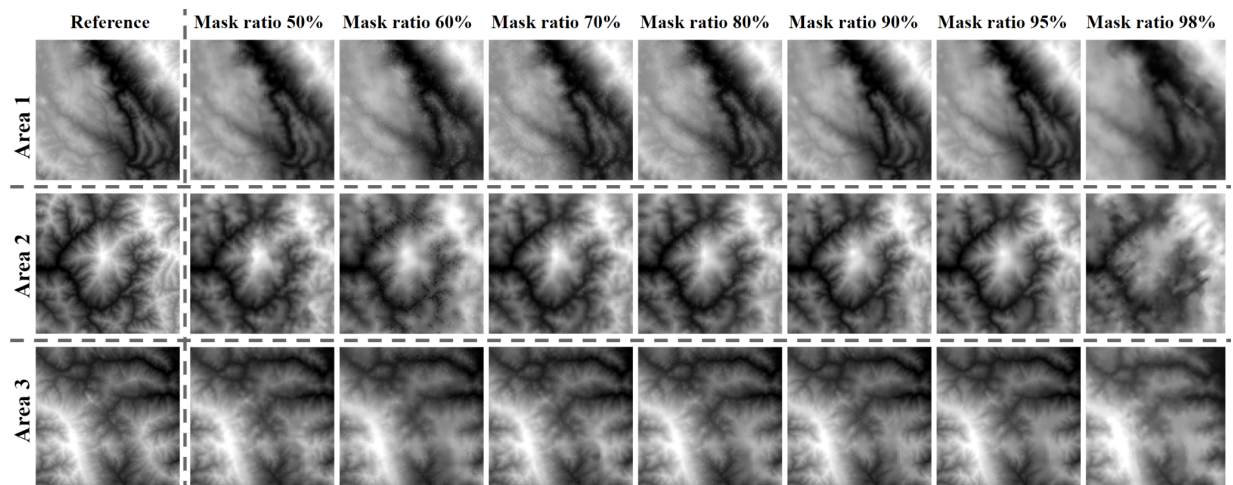

**Fig. 7**. The reference DEM and resultant DEMs of three sub areas interpolated using airborne lidar measurements by the proposed model trained with different mask ratios (from left to right). The size of these sub areas is ~ 30km$^2$.

The above finding is coherent with studies in training the transformer based generative models in different application domains. In the context of self-supervised learning and more specifically in the design of masked autoencoders, a larger mask ratio implies that a greater portion of the input data is obscured or "masked" during the training process. By masking out a larger portion of the input data, the encoder is forced to learn more robust features of the unmasked data in order to reconstruct the original input (He et al., 2021; Xie et al., 2021). This means that instead of relying on superficial or redundant information, the model must identify and utilize the underlying patterns and structures that are truly representative of the data (Xie et al., 2021). In tasks like natural language processing (NLP), a higher mask ratio in training forces the model to understand the context and semantics of the data better, as it cannot rely on nearby words to predict the masked words (Vaswani et al., 2017). Similarly, in image or DEM processing, the model learns to infer the missing parts based on its understanding of the visual context. It is important to note, however, that there is a trade-off. If the mask ratio is too high, the model may not be trained to be able to infer any unseen information at all, which is the situation that the T-GMSI trained with 98% mask ratio shown in Table 4 and Fig. 7. Though it is known that a larger mask ratio leads to better performance, the optimal mask ratio is task dependent. For NLP task, the optimal mask ratio is typically 15% (Devlin et al., 2018), and for computer vision tasks the mask ratio usually ranges from 20% to 75% (He et al., 2021). Therefore, finding the optimal mask ratio is a crucial task when intending to train the proposed model on other interested research.

## 4.2 Effect of sparsity measurements for prediction

Thus far, we have demonstrated the interpolation performance of the proposed model when the large sparsity measurements follow only one specific pattern, i.e., the footprints of space lidar measurements whose sparsity level is at about 92%. To understand the performance of the model under general sparsity pattens, this section will create completely random distributed measurements at different sparsity levels as input to the pre-trained T-GMSI95. To this end, we randomly choose 10,000 30 m 3DEP DEM tiles from the Mendocino County, each of which is 32✕32 grids. Then, different sparsity levels of the measurements are created by random sampling these DEM tiles, as shown in Fig. 8. These randomly distributed measurements are then used to interpolate the DEMs by using the OK, NN, and the pre-trained T-GMSI95 which is proven to be the one with the best performance. The interpolated 30 m DEM tiles are then compared to the reference DEM grid-by-grid. Fig. 8 below gives an example of a DEM tile and the randomly selected measurements to be interpolated with different sparsity levels ranging from 30% to 95%.

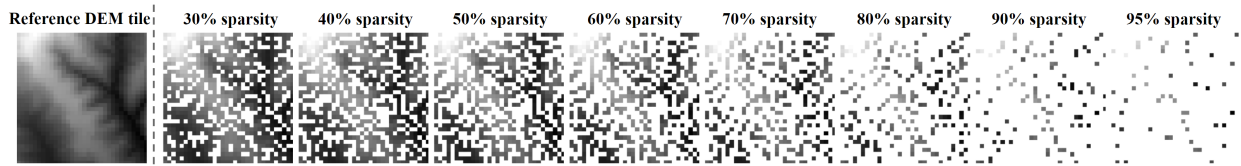

**Fig. 8**. Demonstration of a 32✕32 grids DEM tile and its randomly selected measurements to be interpolated at different sparsity levels ranging from 30% to 95%.

Fig. 9 below presents a series of line graphs, each depicting how different interpolation methods perform in terms of elevation errors as a function of the sparsity level (from 30% to 90%) of randomly distributed measurements. All metrics shown in the figure (average elevation error, standard deviation, mean absolute error, and root mean square error) demonstrate the same pattern: the pre-trained model maintains superior performance and stability across all sparsity levels, particularly excelling beyond 70% sparsity, whereas OK and NN perform well only at lower sparsity levels but deteriorate significantly beyond 70%.

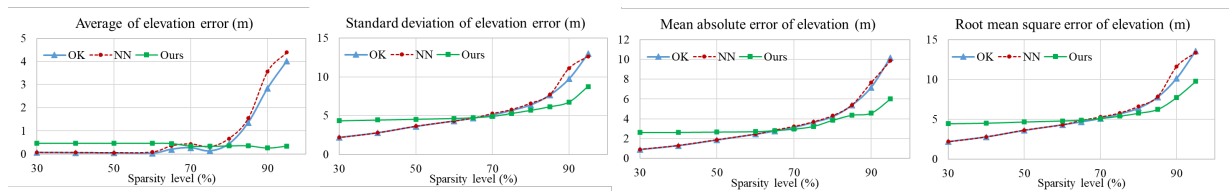

**Fig. 9**. Evaluation of the elevation errors of different methods on interpolation from random measurements at different sparsity levels.

In summary, it can be concluded from Fig. 9 that the pre-trained T-GMSI95 is rather effective and superior in its handling of randomly distributed measurements of high sparsity (>70%), where the absence of data begins to significantly impact the accuracy of the classical interpolation methods. However, when the sparsity level of the data is smaller than 70%, the classical interpolation methods would be a better choice instead; and the performance of the two classical interpolation methods (OK and NN) becomes equivalent as the sparsity level of measurements decreases.

## 4.3 Model transferability

Presuming that our model is pretrained on a wide range of terrains in Mendocino County and has adequately grasped the deep features of the topography, we seek to explore its capability to apply this spatial knowledge to unseen areas with similar or dissimilar landscapes, i.e., the transferability of the model. To this end, we utilize our pretrained T-GMSI95 to Piute County, Tippecanoe County, and Sanborn County. Should the previously acquired deep spatial features be effective in characterizing new geographic patterns, it is anticipated that the proposed model will deliver commendable interpolation outcomes in these unseen areas without the need for any fine tuning to its parameters.

Fig. 10 displays a series of interpolated 30 m DEMs to illustrate the elevation assessment over three selected sub areas in Piute County, Sanborn County, and Tippecanoe County, using the airborne lidar measurements. Visually, compared to the reference 30 m 3DEP DEM, the interpolated DEMs using our pre-trained model display the most accurate representation of the terrain, while the other DEMs from conventional interpolation methods are over smoothed and lack detail structure of the topography. Beyond that, Table 5 summarizes the quantitative evaluation of the elevation difference for the mentioned three counties for transferability investigation using both airborne and spaceborne lidar measurements. When it comes to the improvement achieved by our model, for rugged areas like Piute County, the RMSE decreases ~30% and 15% for airborne lidar measurements and spaceborne lidar measurements compared to the baseline methods. And such reduction of the RMSE is consistent to the reduction of the RMSE in Mendocino County which are ~31% and ~16% for airborne lidar measurements and spaceborne lidar measurements compared to the baseline methods. For the other two counties, Sanborn County and Tippecanoe County, which have very distinct topography and are relatively flat, the improvements over baseline methods become relative subtle with the RMSE decreases by ~17% and ~9% on airborne lidar measurements and spaceborne lidar measurements, respectively. Though such improvement over these two plain areas is less than the hilly area, the results shown in Fig. 10 and Table 5 are still remarkable.

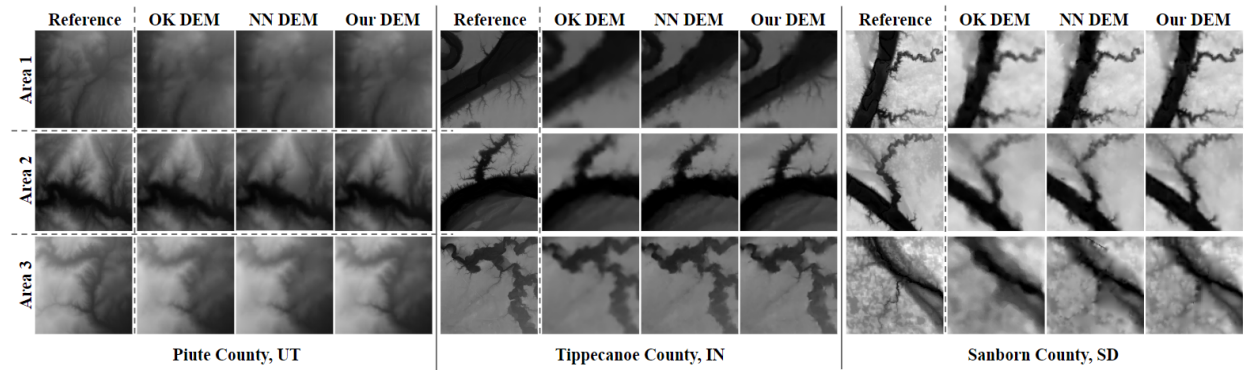

**Fig. 10**. Reference 30 m DEM and interpolated 30 m DEMs using the airborne lidar measurements over three sub areas (from top to bottom) in different counties by different methods (from left to right). The size of every sub area is ~ 30 km$^2$.

**Table 5**. Summary of the elevation assessment in different counties for transferability study, where the T-GMSI model is trained with Mendocino 30m DEM at a mask ratio 95%. The bolded numbers indicate the best results.

|  |  | Ideal/airborne lidar measurements | | | Real/spaceborne lidar measurements | | |
| --- | --- | --- | --- | --- | --- | --- | --- |
|  |  | OK | NN | T-GMSI95 | OK | NN | T-GMSI95 |
| Piute County, UT | $MEAN_{\Delta h}$ (m) | -0.46 | -0.39 | **0.69** | -1.64 | -1.54 | **0.69** |
|  | $STD_{\Delta h}$ (m) | 11.05 | 11.40 | **7.69** | 12.64 | 12.69 | **10.74** |
|  | $MAE_{\Delta h}$ (m) | 5.63 | 5.70 | **4.50** | 6.71 | 6.67 | **6.26** |
|  | $RMSE_{\Delta h}$ (m) | 11.06 | 11.41 | **7.87** | 12.74 | 12.78 | **10.76** |
| Sanborn County, SD | $MEAN_{\Delta h}$ (m) | -0.09 | -0.07 | **0.01** | -0.79 | -0.77 | **-0.68** |
|  | $STD_{\Delta h}$ (m) | 1.21 | 0.96 | **0.87** | 1.27 | 1.09 | **1.06** |
|  | $MAE_{\Delta h}$ (m) | 0.67 | 0.50 | **0.45** | 1.04 | 0.96 | **0.89** |
|  | $RMSE_{\Delta h}$ (m) | 1.21 | 0.96 | **0.88** | 1.50 | 1.33 | **1.27** |
| Tippecanoe County, IN | $MEAN_{\Delta h}$ (m) | 0.06 | -0.12 | **0.02** | -0.40 | -0.35 | **-0.20** |
|  | $STD_{\Delta h}$ (m) | 2.59 | 2.61 | **2.18** | 2.93 | 2.91 | **2.70** |
|  | $MAE_{\Delta h}$ (m) | 1.79 | 1.12 | **1.02** | 1.57 | 1.50 | **1.37** |
|  | $RMSE_{\Delta h}$ (m) | 2.59 | 2.62 | **2.18** | 2.96 | 2.94 | **2.71** |

The observed more obvious improvement of the pre-trained T-GMSI95 in another hilly area compared to plain areas can be attributed to factors that are inherent to the nature of the terrain and the characteristics of the proposed model. The model is learned to recognize and predict patterns based on the features present in the training data. Since the model is trained on data from hilly areas, it should have learned the specific patterns, textures, and elevation gradients characteristic of such terrains. When applied to a similar hilly area, the model can more readily apply these learned features, leading to better performance and more accurate predictions, whereas the conventional methods are unable to use the information of the training sites. The smaller RMSE of the resultant DEM from the T-GMSI in Piute County reflects its ability to capture deep terrain features and its transferability. In addition, it can be inferred that datasets with diverse topographic features and elevation ranges shall be selected as training areas, since this will have significant and profound positive impact on the transferability of the trained model across diverse landscapes.

### 4.4 Comparison with existing CGAN-based method

Beside the conventional interpolation methods such as OK and NN interpolation methods, there are several works focusing on the development of CGAN-based spatial interpolation. The CEDGAN is reported to have decent results on DEM interpolation without the requirements of any additional input other than the sparse terrain measurements (Zhu et al., 2020). In this section, we collect the pre-trained CEDGAN model, which was trained using uniform masking with 90% mask ratio on 32✕32 grids DEM tiles (Zhu et al., 2020), and 33 test DEM tiles provided by the authors of the CEDGAN from the website (https://github.com/dizhu-gis/cedgan-interpolation). The application of the trained CEDGAN model has its limitations (Zhu et al., 2020). (1) A trained CEDGAN model using uniform masking strategy at specific mask ratio would have the best performance on the masked DEM with uniform masking at the same specific sparsity level; (2) A trained CEDGAN model requires a few finetuning before it is applied to another area.

Therefore, instead of using the DEM tiles in our study areas, the collected 33 test DEM tiles will be used to perform the interpolation and evaluation. In addition, uniform masking with 90% mask ratio will be used to mask the collected DEM tiles and generate the masked DEM tiles with 90% sparsity to be interpolated. Then the interpolation on the masked DEM tiles is conducted using the collected pre-trained CEDGAN model and our pre-trained T-GMSI95 based on Mendocino County without any fine tuning. The interpolated DEMs are evaluated with reference to the unmasked test DEM tiles grid by grid.

The RMSE of the interpolated DEMs from the masked DEMs at 90% sparsity level are 2.09 m for our pre-trained model without any fine tuning and 2.60 m (which is consistent with the reported RMSE of 2.587 m in (Zhu et al., 2020) for the pre-trained CEDGAN model. The results are stunning that our pre-trained T-GMSI95 shows ~20% better performance even though no fine tuning has been conducted at all. This finding can be further proven from Fig. 11, which presents the comparison between the interpolated DEM tiles using the pre-trained CEDGAN and pre-trained T-GMSI95. Seven example DEM tiles show that our pre-trained T-GMSI95 is able to replica the majority of the details of the terrain in the reference tiles, and it reconstructs a more similar detailed topography to the reference DEM tiles than the CEDGAN model.

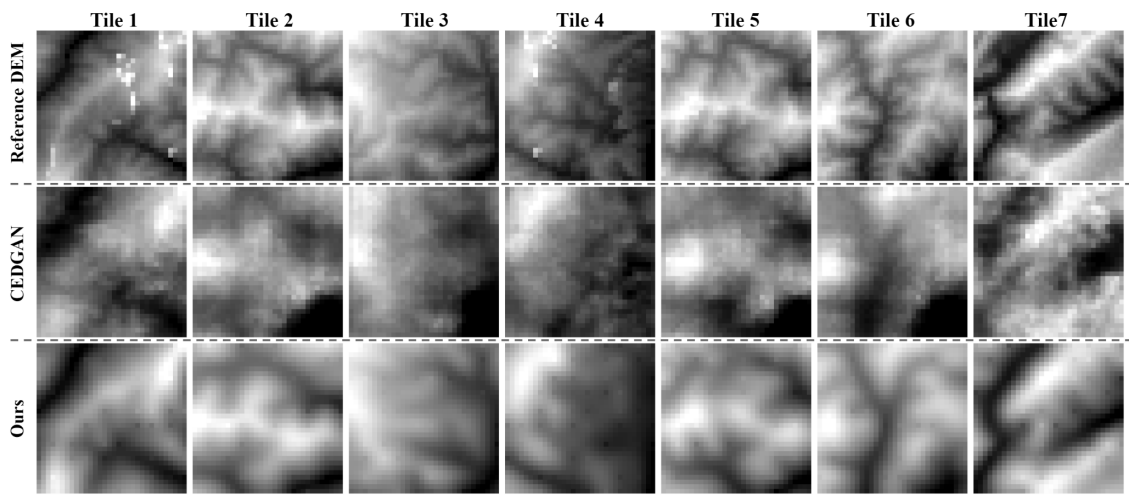

**Fig. 11**. Reference DEM tiles and interpolated DEM tiles using the CEDGAN model and our model (from top to bottom). Each of the DEM tiles consists of 32×32 grids.

## 5. Conclusions

This paper introduces T-GMSI, a spatial interpolator using transformer-based encoder-decoder generative model for large sparsity data. While applicable across various environmental fields, the model's efficacy is demonstrated using DEM data in this study. The innovative aspects of the proposed model are threefold. Firstly, it incorporates a ViT-based encoder for deep feature extraction and a ViT-based decoder for spatial interpolation. Secondly, the model's loss module integrates a conventional reconstruction loss function with a topography-aware gradient loss function. Thirdly, a pre-trained model, using a high mask ratio (95%) during the training procedure, exhibits notable transferability to different topography without necessitating additional fine-tuning in different study areas other than the training area.

Extensive empirical evaluations affirm the T-GMSI's effectiveness and transferability. First of all, experiments show that the proposed model obtains the best performance when trained with a random masking strategy at 95% mask ratio. Moreover, comparative analysis reveals that T-GMSI outperforms existing interpolation methods in terms of accuracy and visual quality on DEMs derived from large sparsity measurements. Specifically, relative to OK and NN interpolation methods, T-GMSI reduces the RMSE by 40% and 25% respectively on airborne lidar measurements, and by 23% and 10% on space lidar measurements containing larger elevation measurement errors. The transferability of the pre-trained T-GMSI is corroborated through experiments across three diverse U.S. counties. Compared to OK and NN methods, the RMSE improvement of T-GMSI is 29% and 31% for counties with similar landscapes to the training dataset, and 16%-17% and 8%-27% for counties with differing topographies. Additionally, compared to the reported CGAN-based model (CEDGAN), T-GMSI improves RMSE by 20% on CEDGAN-provided DEM data without any fine-tuning. Results demonstrate that T-GMSI maintains high performance across varying landscapes for measurements with large (>70%) sparsity levels.

In conclusion, T-GMSI not only excels in generating high-quality elevation surfaces from large sparsity measurements but also achieves SOTA performance, illustrating the necessity of developing deep learning methods specifically tailored to geographical data instead of direct adoption of common deep learning approaches from information science. It should be pointed out that the proposed T-GMSI can be regarded as a paradigm for spatial interpolation not only for terrain modeling but also other kinds of geospatial applications. Its effectiveness and limitations require further exploration by research in other applied environmental modeling context.

to predict soil heavy metals based on a genetic algorithm and neural network model. Sci. Total Environ. 825, 153948.
Zhang, W., Liu, Y., Dong, C., Qiao, Y., 2019. RankSRGAN: Generative adversarial networks with ranker for image super-resolution, in: 2019 IEEE/CVF International Conference on Computer Vision (ICCV). Presented at the 2019 IEEE/CVF International Conference on Computer Vision (ICCV), IEEE, pp. 3096–3105.
Zhang, Y., Yu, W., 2022. Comparison of DEM Super-Resolution Methods Based on Interpolation and Neural Networks. Sensors 22. https://doi.org/10.3390/s22030745
Zhang, Y., Yu, W., Zhu, D., 2022. Terrain feature-aware deep learning network for digital elevation model superresolution. ISPRS J. Photogramm. Remote Sens. 189, 143–162.
Zhou, T., Li, Q., Lu, H., Cheng, Q., Zhang, X., 2023. GAN review: Models and medical image fusion applications. Inf. Fusion 91, 134–148.
Zhu, D., Cheng, X., Zhang, F., Yao, X., Gao, Y., Liu, Y., 2020. Spatial interpolation using conditional generative adversarial neural networks. Int. J. Geogr. Inf. Sci. 34, 735–758.24